# Co-Linguistics:
# AI-augmented Theory Construction in Linguistics

Emmanuel Chemla, Benjamin Spector, Alexandros Kalomoiros, Philippe Schlenker

**Abstract.** LLMs have been studied in recent linguistics as potential models of humans' linguistic abilities. Here we discuss an entirely different use of AI, namely as a co-scientist, to help construct and assess linguistic theories (we refer to the result as "Co-Linguistics"). Since the 1960s, linguistics has developed theories that are in principle mathematically formalizable, often in the language of formal language theory or model theory. The AI revolution in mathematics will thus have consequences in linguistics—but with an essential twist: proving new theorems is rarely the linguist's goal. Rather, one seeks to find the best set of axioms to derive empirical statements. AI could accelerate research by making existing theories fully explicit, by comparing competing theories, and more ambitiously, by proposing new theories (in machine learning, this relates to "program induction"). It will also help assess theories by accelerating the identification and test of crucial predictions, thanks to unparalleled access to data (in machine learning, this relates to "active learning"). While the cycle from theory evaluation to theory construction may give rise to recursive and possibly autonomous improvement of linguistic theories, humans remain central: linguists provide scientific directions and evaluate theories conceptually, and experimental participants are needed to assess empirical predictions that are outside the reach of LLMs.

# 1. Introduction

## 1.1 Goals

There has recently been considerable research, and debates, about the use of LLMs as models of human linguistic abilities (Choshen et al 2026, Vázquez Martínez & Yang 2026). There has also been important research on their (possibly non-human like) linguistic abilities (e.g., Wilcox et al 2018, Yedetore et al 2023), and even of their metalinguistic abilities (Beguš et al 2025a). Here we discuss a different use of AI for linguistics, namely as a co-scientist, to help construct and assess linguistic theories (the result may be called "co-linguistics", i.e. the construction of linguistic theories with the help of an AI co-worker). While we think this tool will help advance linguistic theory, it is a separate question whether it will be good for linguists as individuals or for linguistics as a field. This should be the object of a separate debate.

Since its inception around the 1960s, formal linguistics has sought to develop entirely formalizable theories. These aim to be statable in the language of formal language theory, model theory, and other areas of mathematics. In principle, then, linguistic theories are collections of axioms that mechanically derive desirable empirical consequences. In recent months, the AI revolution in mathematics has become obvious: LLMs have made it possible to accelerate the formalization of important areas of mathematics, in ways that are checkable with proof-assistants[1]. More strikingly, LLMs have helped solve some problems that were thought to be very important to contemporary mathematics (e.g., Tao 2026). In other words, machines can generate proofs, not just check them.

[1] It was of course known that mathematics can in principle be formalized (specifically, in the language of set theory - see for instance Bagaria 2023). But it is a different matter to actually formalize proofs mostly written in natural language.

Since linguistic theories are in principle mathematical theories, the AI revolution in mathematics can be expected to have consequences for linguistics, but with an essential twist. Only rarely does the difficulty lie in establishing or proving the derivation. Rather, the name of the game is usually to find the best set of axioms to derive desirable empirical statements (in machine learning, this relates to “program induction”).

We argue that AI could accelerate linguistic research by making existing theories fully explicit, by comparing competing theories, and more ambitiously, by proposing new theories. It will also help assess theories by accelerating the identification and test of crucial predictions (in machine learning, this relates to “active learning”). For the sake of concreteness, we provide at multiple junctures illustrations from our own experience in semantics, but we hope that our general message applies across subfields of linguistics.[2] Details about our case studies are found in the online Appendix, whose content will change as we obtain sharper results.

While AI may well accelerate the recursive improvement of linguistic theories, from theory evaluation to theory construction, as illustrated in Figure 1, humans remain central: linguists have to provide scientific directions and evaluate theories (as the ultimate goal is human understanding), and introspective judgments and/or more involved psycholinguistic experiments are needed to assess empirical predictions that are likely outside the reach of LLMs.

The co-linguistic loop:
AI-Augmented Theory Construction
1. Theory Construction
Human-based and/or LLM-based definition of principles
2. Formal Verification
Lean formalization of the principles & verification of the derivation of their consequences
3. Evaluation of explanatory power
Minimum Description Length
Generality of primitives
Human assessment
4. Empirical evaluation
Introspective judgments
Psycholinguistic experiments
Corpus data & mining
Active learning queries
Recursive improvement loop. New empirical findings testing crucial predictions (Step 4) feed back into theory construction through the exploration of new principles (Step 1), thus accelerating the cycle of scientific discovery.

**Figure 1:** AI could be used at various stages of the scientific process and accelerate the recursive improvement loop, in science in general and in linguistics in particular.

## 1.2 Precedents and parallels

Our proposal has several precedents, and there are parallel proposals underway in other fields.

The use of proof-assistants was advocated in semantics by Chatzikyriakidis and Luo (2016), who explicitly proposed that this made it possible to “experiment on new theories”. More recently, Chojecki (2026) goes one step further and proposes to use the proof-assistant Lean for science formalization across fields, such as biology, chemistry and economics, with an explicit proposal that untrusted theory generators should be coupled with trusted proof-assistants, giving rise to a recursive loop. This is the very model we discuss for linguistics. Watchus (2025) and Ma et al (2025) propose related architectures for physics.

[2] We focus on our own work because we know it well, not because it is especially important.

As things stand, proof-assistants appear to be used on a large scale in mathematics, theoretical physics, and economics.[3] Once proof-assistants are widely used in a field, it is natural to couple them with generative AI to come up with new proofs and theories. On the technological front, new varieties of AI for science seem to pop up almost every day, with the kind of recursive improvement loop we discussed above (the most recent at the time of writing comes from Xue et al 2026).

Special mention should be made of Bayesian Program Induction, which searches through a space of possible programs (stated in a predetermined formal language) with the goal of favoring program simplicity and predictive accuracy. This can be seen as a procedure to discover new theories. As an example within linguistics, Ellis et al (2022) seek to automate theory-construction for morpho-phonology, with the dual goal of modeling the child's and the scientist's discovery procedure. This proposal is both more narrow and more ambitious than recent recipes to use generative AI coupled with proof-assistants. In Ellis et al's proposal, the format of rules and the way to explore the space of possible theories is predefined. The model aims to shed light on child development, under the assumption that the child's acquisition procedure is in some sense optimal (Bayesian), and based on the same predefined format of rules. By contrast, in recent AI-driven recipes, the format of theories is only constrained by the relevant AI's "imagination" (hence indirectly by its training set and internal biases), and the search procedure is similarly constrained by whatever heuristics the AI may think of, possibly augmented with human intervention. As a way to model the child's discovery procedure, this is of course a non-starter, unlike Ellis et al's ambitious model.

# 2. AI-assisted Theory Development

From phonology to morphology, syntax, semantics and pragmatics, formal linguistics has sought to develop highly explicit theories of human linguistic behavior, ones that can in principle be expressed within formal language theory, model theory or other areas of mathematics (linguistic methods can also be used to investigate the communicative abilities of different entities, such as apes, monkeys, birds and bees, or for that matter LLMs themselves; we come back to these extensions below).

Inevitably, then, the successes of AI in mathematics have immediate consequences for linguistics, of three types. First, AI can help formalize theories and make them fully explicit—the original promise of formal linguistics. Second, AI can help compare formal theories, proving equivalences and non-equivalences, and extracting crucial differences in predictions. Third, and most ambitiously, AI can help construct new theories, either by tweaking existing ones or by exploring more radically different sets of principles.

## 2.1 Formalizing existing theories

While linguistic theories rarely give rise to sophisticated theorems, fully formalizing such theories isn't trivial and can help clarify theoretical claims, uncover gaps, and check formal results efficiently. Similar points have been made across fields that use formalization and now proof-assistants (e.g. Chojecki 2026).

To illustrate, we asked Claude to re-formalize the geometric part of Schlenker and Lamberton (2024), which provided an explicit semantics for highly iconic constructions of sign language, called “classifiers”. Claude offered two equivalent restatements in cleaner and simpler geometric terms, and offered verifications of the equivalences using the proof-assistant Lean.

[3] For the use of Lean in mathematics, see https://leanprover-community.github.io/papers/mathlib-paper.pdf; for theoretical physics, see Tooby-Smith (2025); for economics, see Garg (2026).

In a more traditional semantic task, we asked Claude to derive the predictions of a theory of presupposition projection, that of Kalomoiros (2022, 2023), for the case of sentences with quantifiers. Kalomoiros himself had not discussed this case within his “Limited Symmetry” framework. Sample predictions were derived by hand in Schlenker (2025). Claude re-proved these predictions and derived further facts for additional quantifiers.

As mentioned earlier, a key step in verifying proofs is the use of a proof-assistant—here, Lean 4 (de Moura & Ullrich 2021). This removes the challenge of ensuring that the steps are correct, but this emphatically does not obviate the need to check that the principles have been formalized correctly and that the theorems correspond to what one had in mind. AI can help with the formalization step, but in the end linguists are responsible for checking that it is adequate. Correspondingly, our results in the Appendix are provisional: more work is needed to check that the AI formalization matches what the original theories intended.

## 2.2 Comparing existing theories

Once theories are formalized, one can seek to prove equivalence or non-equivalence results among them, and to isolate crucial predictions that may adjudicate among them.

We did this in a simple case for our running example involving presuppositions. First, we asked Claude to check some equivalence results among five theories of presupposition projection that were originally proven “by hand” in an unpublished addition to Schlenker (2009). As outlined in the Appendix, we used Claude to check and correct the results, and also propose new results that were not in the original piece. Second, in the more recent case of Kalomoiros’s “Limited Symmetry” theory, Claude offered general theorems connecting it to earlier analyses (the Transparency/Local Contexts systems of Schlenker 2008, 2009), thus shedding new light on theory comparison.

## 2.3 Building new theories

The key promise of AI-augmented theory construction is to explore *new* theories that can compete with, and eventually improve on, existing ones. Initially one might want to explore tweaks to existing theories, but things will get more interesting when one uses AI to suggest genuinely new theoretical ideas.[4]

As a proof of concept, consider anaphoric licensing. Pronouns are often licensed by a preceding indefinite, as in *There is [a bathroom]_i and it_i is upstairs*, or *There isn’t [a bathroom]_i or it is upstairs.* Building on the literature, Spector (2026) proposes a new theory of anaphora (among many others) which aims to predict the licensing and interpretation of singular pronouns in virtually any environment. The theory faces some limitations. In particular, it only allows for *left-to-right* licensing, which is appropriate for conjunction (hence the deviance of: *It_i is upstairs and there is [a bathroom]_i) but not for disjunction, which allows for cataphora (Either it_i is upstairs, or there isn’t [a bathroom]_i). Revising the theory to solve this problem while preserving the positive predictions of the theory proved challenging. We used OpenAI’s Codex to improve (part of) Spector’s proposal. The result builds on Kalomoiros (2022, 2023) to solve the cataphora problem. In this case, then, the linguist and AI co-linguist developed a new theory of anaphora, which seems no more complex than Spector’s original theory, but has better empirical coverage.

[4] Relatedly, in a project on sperm whale sequences, Beguš et al (2025b) find counterparts of vowels and diphthongs with the help of a deep neural network architecture (fiwGAN, Beguš 2021).

# 3. AI-assisted Theory Evaluation

Theory evaluation can take three forms, which we discuss in turn below.

§3.1. Keeping the predictions constant, are there differences of explanatory power, perspicuousness and elegance among competing theories?

§3.2. In view of all the published data in linguistics, what are the strengths and weaknesses of competing theories?

§3.3. Can one collect new data to adjudicate among competing theories?

## 3.1 How explanatory is the theory?

There is currently no automatic procedure to assess the explanatory value of a theory: scientists are the ultimate judges, because the goal of science is human understanding. But AI can help assess some important criteria.

1. Parsimony can be evaluated by the length of a theory, operationalized by way of Minimum Description Length (MDL): all else being equal, a theory that can be expressed more compactly is preferable. This is necessarily relative to an implementation language (e.g. Lean vs. Python). In practice, code length can be a useful initial approximation of parsimony.
2. The generality of a theory is a criterion as well: all else being equal, principles that are applicable to more data are preferable. For instance, it is preferable for a linguistic principle to apply to several languages than to just one.
3. Further criteria could be proposed, and assessed “by hand” or with the help of AI. In addition, to the extent that human specialists remain the ultimate judges of explanatory value, LLMs could be used as an initial proxy of human judgment—with the obvious risk that this proxy might be wrong.

## 3.2 How empirically adequate is the theory?

As linguistics is usually conceived, its object of study is human language. But linguistic-like methods can be used to investigate animal communication, as in recent animal linguistics (e.g. Schlenker et al 2016, Suzuki 2021, Berthet et al 2023) or the linguistic abilities of LLMs themselves (Choshen et al 2026, Vázquez Martínez & Yang 2026). Here we discuss potential applications and limitations of AI for gathering data across these three different, but related, areas.

Starting with the human case, it is clear that AI is a powerful tool to collect published results in linguistics, and to bring them to bear on crucial predictions. Various online tools have been proposed to conduct literature review and meta-analyses. Critical examples can be semi-automatically searched and extracted from them, and important ongoing initiatives seek to organize this large-scale empirical work (from Gauthier et al 2020 to Ying et al 2026).

AI may also help collect new data. If the target data involve corpora, LLMs will have immediate benefits (because of their huge training set, and access to additional data online). Often, however, crucial predictions do not involve what is exemplified in texts, but more complex data pertaining to comparative acceptability, inferential judgments, truth-conditional judgments as well as processing times, or data about child development and brain localization, for instance. These motivate the use of detailed introspective judgments in theoretical linguistics, and of sophisticated experiments in psycholinguistics.

Here one may consider two kinds of shortcuts. One is to take LLMs’ behavior as a proxy for human behavior, and the other is to ask LLMs to predict the results of crucial psycholinguistic experiments. These proxies offer a fast track to results, but they are obviously error-prone. On the other hand, one might have a prior assessment of LLMs' track-record in making such predictions in a given area (e.g., we might have learned that LLMs are good at predicting a certain kind of psycholinguistic judgment). This could then provide a modicum of confidence in such predictions. Theoretical iteration can proceed while keeping track of each proxy's precise reliability (for the risks involved, see Qiu, Duan & Cai 2023, Trott 2024, Qiu, Duan & Cai 2025, Hwang 2026, among others).

It is clear that, in the end, human linguistic judgments will be needed to adjudicate among competing theories. This is especially true because critical judgments often involve sophisticated examples that are at the margins of what humans actually use, but nevertheless yield quite uniform intuitions—much as theories of perception are informed by the study of rare illusions. Such crucial data are likely to lie outside the comfort zone of the training set of LLMs.

When the object of study is animal communication rather than human language, data are typically very limited, and testing new predictions usually requires setting up new experiments. The increasing availability of large multimodal corpora of animal behavior may soon change things, however.

When the object of study is the linguistic behavior of LLMs themselves, we are in the opposite situation: new data may be collected in an automatic fashion. Data may be linguistic in nature—e.g. new sentences that test crucial predictions about LLM behavior. But they may also be the LLM counterparts of neuroscientific data. Their accessibility makes it easy to conduct experiments on LLMs that would require the most intrusive techniques if applied to the human brain (Lakretz et al 2019, Lakretz et al 2021, Bereska & Gavves 2024).

# 4. Recursive Improvement of Theories

The AI contributions we have sketched are modular. AI may in particular: 1. Construct multiple new theories. 2. Check that conclusions follow from principles—namely as theorems derived from axioms. 3. Assess competing theories on the basis of their relative parsimony. 4. Find new data to adjudicate among competing theories.

Importantly, when all four components can be effected with AI, theories may be recursively improved, with new data (Step 4) feeding into theory construction (Step 1) (see Figure 1). But humans obviously remain central (to put it in AI terms, we advocate a "human in the loop" procedure). First, human experts must act as judges, as they are best-positioned to assess the scientific usefulness of a theory: they must filter out theories and guide further iterations. Second, for predictions about human linguistics abilities, there is no escape from collecting data from humans—and similarly with animals when the data pertain to animal communication. By contrast, when the target is the linguistic behavior of LLMs, this empirical part of recursive improvement of theories may be fully automated.

In all cases, AI can be expected to significantly accelerate theoretical progress.

# 5. Conclusion

AI could prove to be a powerful tool to formalize, verify, and even construct theories in linguistics. It could help assess them, both in terms of parsimony and empirical coverage, including by finding new data to test crucial predictions. When all steps are present, AI could give rise to recursive improvement of theories. AI thus has a huge potential to improve the depth, breadth and speed of theory construction in linguistics.

As we highlighted at the outset, it does not follow that this development will be good for linguists as individuals, or for linguistics as a field. This is a separate issue, which should be the object of a broad debate.

# Appendix

Our ongoing case studies (still in flux at the time of writing) can be found at the following link: https://tinyurl.com/colinguistics

# Acknowledgements

Part of this research was conducted at DEC, Ecole Normale Supérieure - PSL Research University. DEC is supported by grant FrontCog ANR-17-EURE-0017. Schlenker gratefully acknowledges the support of the Humboldt Foundation (Humboldt Research Award 2025).